\pdfoutput=1
\documentclass[10pt,twocolumn,letterpaper]{article}

\usepackage[absolute]{textpos}
\usepackage{wacv}
\usepackage{times}
\usepackage{epsfig}
\usepackage{graphicx}
\usepackage{amsmath}
\usepackage{amssymb}

\usepackage{subcaption}
\usepackage{booktabs}
\usepackage{tabularx}
\usepackage{enumitem}
\usepackage{color,xcolor}
\PassOptionsToPackage{hyphens}{url}

\usepackage[pagebackref=true,breaklinks=true,letterpaper=true,colorlinks,bookmarks=false]{hyperref}
\usepackage[capitalize]{cleveref}

\makeatletter
\renewcommand*\subcaption@label{%
    \caption@withoptargs\subcaption@@label}
\makeatother

\newcommand{\cossim}{\sigma_\mathrm{cos}}
\newcommand{\cosloss}{\mathcal{L}_\mathrm{cos}}
\newcommand{\cosclsloss}{\mathcal{L}_\mathrm{cos+xent}}
\newcommand{\xentloss}{\mathcal{L}_\mathrm{xent}}
\newcommand{\mseloss}{\mathcal{L}_\mathrm{MSE}}

\definecolor{fsuBlue}{RGB}{135,178,231}

\wacvfinalcopy 

\def\wacvPaperID{953} 

\ifwacvfinal\fi
\begin{document}

\title{Deep Learning on Small Datasets without Pre-Training using Cosine Loss}

\author{Bj{\"o}rn Barz \hspace{2cm} Joachim Denzler \\
    \\
    Friedrich Schiller University Jena, Germany\\
    Computer Vision Group\\
    {\tt\small bjoern.barz@uni-jena.de}
}

\maketitle
\ifwacvfinal\thispagestyle{empty}\fi

\begin{textblock*}{\textwidth}(18mm,12mm)
    \textblockcolour{fsuBlue}
    \vspace{2mm}
    \tiny
    \centering
    Bj{\"o}rn Barz and Joachim Denzler.\\
    ``Deep Learning on Small Datasets without Pre-Training using Cosine Loss.''\\
    \textit{IEEE Winter Conference on Applications of Computer Vision (WACV) 2020.}\\
    \copyright\ 2020 IEEE. Personal use of this material is permitted.
    Permission from IEEE must be obtained for all other uses, in any current or future media, including reprinting/republishing this material for advertising or promotional purposes, creating new collective works, for resale or redistribution to servers or lists, or reuse of any copyrighted component of this work in other works.
    The final publication will be available at
    \href{https://ieeexplore.ieee.org/}{ieeexplore.ieee.org}.\\
    \vspace{2mm}
\end{textblock*}

\begin{abstract}
Two things seem to be indisputable in the contemporary deep learning discourse:
1.~The categorical cross-entropy loss after softmax activation is the method of choice for classification.
2.~Training a CNN classifier from scratch on small datasets does not work well.

In contrast to this, we show that the cosine loss function provides substantially better performance than cross-entropy on datasets with only a handful of samples per class.
For example, the accuracy achieved on the CUB-200-2011 dataset without pre-training is by 30\% higher than with the cross-entropy loss.
Further experiments on other popular datasets confirm our findings.
Moreover, we demonstrate that integrating prior knowledge in the form of class hierarchies is straightforward with the cosine loss and improves classification performance further.
\end{abstract}

\section{Introduction}
\label{sec:introduction}

\noindent
Deep learning methods are well-known for their demand after huge amounts of data \cite{sun2017revisiting}.
It is even widely acknowledged that the availability of large datasets is one of the main reasons---besides more powerful hardware---for the recent renaissance of deep learning approaches \cite{krizhevsky2012alexnet,sun2017revisiting}.
However, there are plenty of domains and applications where the amount of available training data is limited due to high costs induced by the collection or annotation of suitable data.
In such scenarios, pre-training on similar tasks with large amounts of data such as the ImageNet dataset \cite{deng2009imagenet} has become the de facto standard \cite{zeiler2014visualizing,girshick2014rcnn}, for example in the domain of fine-grained recognition \cite{lin2015bilinear,zheng2017learning,cui2018large,Simon19:implicit}.

While this so-called \textit{transfer learning} often comes without additional costs for research projects thanks to the availability of pre-trained models, it is rather problematic in at least two important scenarios:
On the one hand, the target domain might be highly specialized, e.g., in the field of medical image analysis \cite{litjens2017survey}, inducing a large bias between the source and target domain in a transfer learning scenario.
The input data might have more than three channels provided by sensors different from cameras, e.g., depth sensors, satellites, or MRI scans.
In that case, pre-training on RGB images is anything but straightforward.
But even in the convenient case that the input data consists of RGB images, legal problems arise:
Most large imagery datasets consist of images collected from the web, whose licenses are either unclear or prohibit commercial use \cite{deng2009imagenet,krizhevsky2009cifar,tencent-ml-images-2019}.
Therefore, copyright regulations imposed by many countries make pre-training on ImageNet illegal for commercial applications.
Nevertheless, the majority of research applying deep learning to small datasets focuses on transfer learning.
Given huge amounts of data, even simple models can solve complex tasks by memorizing \cite{torralba2008tinyimages,zhang2016understanding}.
Generalizing well from limited data is hence the hallmark of true intelligence.
But still, works aiming at directly learning from small datasets without external data are surprisingly scarce.

Certainly, the notion of a ``small dataset'' is highly subjective and depends on the task at hand and the diversity of the data, as expressed in, e.g., the number of classes.
In this work, we consider datasets with less than 100 training images per class as small, such as the Caltech-UCSD Birds (CUB) dataset \cite{WahCUB_200_2011}, which comprises at most 30 images per class.
In contrast, the ImageNet Large Scale Visual Recognition Challenge 2012 (ILSVRC'12) \cite{russakovsky2015ilsvrc} contains between 700 and 1,300 images per class.

Since transfer learning works well in cases where sufficiently large and licensable datasets are available for pre-training, research on new methodologies for learning from small data \textit{without external information} has been very limited.
For example, the choice of categorical cross-entropy after a softmax activation as loss function has, to the best of our knowledge, not been questioned.
In this work, however, we propose an extremely simple but surprisingly effective loss function for learning from scratch on small datasets: the \textit{cosine loss}, which maximizes the cosine similarity between the output of the neural network and one-hot vectors indicating the true class.
Our experiments show that this is superior to cross-entropy by a large margin on small datasets.
We attribute this mainly to the $L^2$ normalization involved in the cosine loss, which seems to be a strong, hyper-parameter free regularizer.
\\
In detail, our contributions are the following:
\begin{enumerate}[leftmargin=*]
    \item We conduct a study on 5 small image datasets (CUB, NAB, Stanford Cars, Oxford Flowers, MIT Indoor Scenes) and one text classification dataset (AG News) to assess the benefits of the cosine loss for learning from small data.
    
    \item We analyze the effect of the dataset size using differently sized subsets of CUB, CIFAR-100, and AG News.
    
    \item We investigate whether the integration of prior semantic knowledge about the relationships between classes as recently suggested by Barz and Denzler  \cite{Barz19:Embeddings} improves the performance further.
    To this end, we introduce a novel class taxonomy for the CUB dataset and also evaluate different variants to analyze the effect of the granularity of the hierarchy.
\end{enumerate}

\noindent
The remainder of this paper is organized as follows:
We will first briefly discuss related work in \cref{sec:related-work}.
In \cref{sec:cosine-loss}, we introduce the cosine loss function and briefly review semantic embeddings \cite{Barz19:Embeddings}.
\Cref{sec:datasets} follows with an introduction of the datasets used for our empirical study in \cref{sec:experiments}.
A summary of our findings in \cref{sec:conclusions} concludes this work.

\begin{figure*}[t]
    \includegraphics[width=\linewidth]{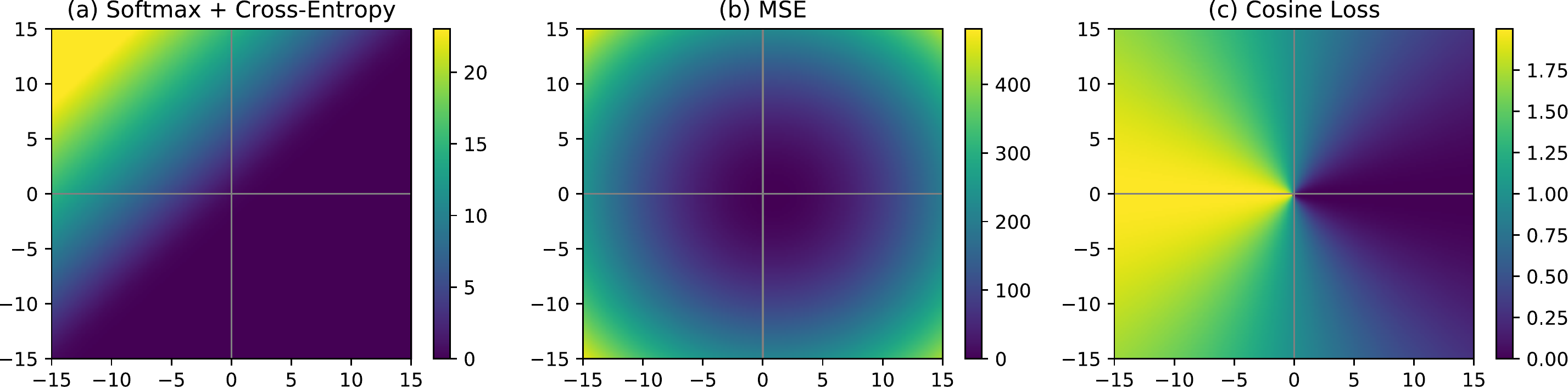}%
    {\phantomsubcaption\label{subfig:xent-loss}}%
    {\phantomsubcaption\label{subfig:mse-loss}}%
    {\phantomsubcaption\label{subfig:cosine-loss}}%
    \caption{Heatmaps of three loss functions in a 2-D feature space with fixed target $\varphi(y)=\protect\begin{bmatrix}1&0\protect\end{bmatrix}^\top$.}
    \label{fig:loss-functions}
\end{figure*}

\section{Related Work}
\label{sec:related-work}

\paragraph{Learning from Small Data}

The problem of learning from limited data has been approached from various directions.
First and foremost, there is a huge body of work in the field of \textit{few-shot and one-shot learning}.
In this area, it is often assumed to be given a set of classes with sufficient training data that is used to improve the performance on another set of classes with very few labeled examples.
\textit{Metric learning} techniques are common in this scenario, which aim at learning discriminative features from a large dataset that generalize well to new classes \cite{vinyals2016matching,snell2017prototypical,sung2018learning,wang2018cosface,wu2018improving}, so that classification in face of limited data can be performed with a nearest neighbor search.
Another approach to few-shot learning is \textit{meta-learning}: training a learner on large datasets to learn from small ones \cite{li2017meta,qiao2018few,wang2019tafe}.

Our work is different from these few-shot learning approaches due to two reasons:
First, we aim at learning a deep classifier entirely from scratch on small datasets, without pre-training on any additional data.
Secondly, our approach covers datasets with roughly between 20 and 100 samples per class, which is in the interstice between a typical few-shot scenario with even fewer samples and a classical deep learning setting with much more data.

Other approaches on learning from small datasets employ domain-specific \textit{prior knowledge} to either artificially enlarge the amount of training data or to guide the learning.
Regarding the former, Hu et al.\ \cite{hu2018frankenstein}, for instance, composite face parts from different images to create new face images and Shrivastava et al.\ \cite{shrivastava2017learning} conduct training on both real images and synthetic images using a GAN.
As an example for integrating prior knowledge, Lake et al.\ \cite{lake2015human} represent classes of handwritten characters as probabilistic programs that compose characters out of individual strokes and can be learned from a single example.
However, generalizing this technique to other types of data is not straightforward.

In contrast to all approaches mentioned above, our work focuses on learning from limited amounts of data without any external data or prior knowledge.
This problem has recently also been tackled by incorporating a GAN for data augmentation into the learning process \cite{zhang2018dada}.
As opposed to this, we approach the problem from the perspective of the loss function, which has not been explored extensively so far for direct fully-supervised classification.

\paragraph{Cosine Loss}

The cosine loss has already successfully been used for applications other than classification.
Qin et al.\ \cite{qin2008rankcosine}, for example, use it for a list-wise learning to rank approach, where a vector of predicted ranking scores is compared to a vector of ground-truth scores using the cosine similarity.
It furthermore enjoys popularity in the area of cross-modal embeddings, where different representations of the same entity, such as images and text, should be close together in a joint embedding space \cite{sudholt2017evaluating,salvador2017learning}.

Various alternatives for the predominant cross-entropy loss have furthermore recently been explored in the field of deep metric learning, mainly in the context of face identification and verification.
Liu et al.\ \cite{liu2017sphereface}, for example, extend the cross-entropy loss by enforcing a pre-defined margin between the angle of features predicted for different classes.
Ranjan et al.\ \cite{ranjan2017l2}, in contrast, $L^2$-normalize the predicted features before applying the softmax activation and the cross-entropy loss.
However, they found that doing so requires scaling the normalized features by a carefully tuned constant to achieve convergence.
Wang et al.\ \cite{wang2018cosface} combine both approaches by normalizing both the features and the weights of the classification layer, which realizes a comparison between the predicted features and learned class-prototypes by means of the cosine similarity.
They then enforce a margin between classes in angular space.
While such techniques are also sometimes referred to as ``cosine loss'' in the face identification literature, the actual classification is still performed using a softmax activation and supervised by the cross-entropy loss, whereas we use the cosine similarity directly as a loss function.
Furthermore, the aforementioned methods focus rather on learning image representations that generalize well to novel classes (such as unseen persons) than on directly improving the classification performance on the set of classes on which the network is trained.
Moreover, they introduce new hyper-parameters that must be tuned carefully.

In the context of image retrieval, Barz and Denzler \cite{Barz19:Embeddings} recently used the cosine loss to map images onto semantic class embeddings derived from a hierarchy of classes.
While the focus of their work was to improve the semantic consistency of image retrieval results, they also reported classification accuracies and achieved remarkable results on the NAB dataset without pre-training.
This led them to the hypothesis that prior semantic knowledge would be particularly useful for fine-grained classification tasks.
In this work, we show that the main reason for this phenomenon is the use of the cosine loss and that it can be applied to {\emph any} small dataset to achieve better classification accuracy than with the standard cross-entropy loss, even with one-hot vectors as class embeddings.
The effect of semantic class embeddings, in contrast, is only complementary.

\section{Cosine Loss}
\label{sec:cosine-loss}

\noindent
In this section, we introduce the cosine loss and briefly review the idea of hierarchy-based semantic embeddings \cite{Barz19:Embeddings} for combining this loss function with prior knowledge.

\subsection{Cosine Loss}
\label{subsec:cosine-loss}

\noindent
The \textit{cosine similarity} between two $d$-dimensional vectors $a, b \in \mathbb{R}^d$ is based on the angle between these two vectors and defined as
\begin{equation}
\label{eq:cosine-similarity}
\cossim(a, b) = \cos(a \angle b) = \frac{\langle a, b \rangle}{\|a\|_2 \cdot \|b\|_2} \;,
\end{equation}
where $\langle \cdot, \cdot \rangle$ denotes the dot product and $\|\cdot\|_p$ the $L^p$ norm.

Let $x \in \mathfrak{X}$ be an instance from some domain (images, text etc.) and $y \in \mathcal{C}$ be the class label of $x$ from the set of classes $\mathcal{C} = \{1, \dotsc, n\}$.
Furthermore, $f_\theta: \mathfrak{X} \rightarrow \mathbb{R}^d$ denotes a transformation with learned parameters $\theta$ from the \textit{input space} $\mathfrak{X}$ into a $d$-dimensional Euclidean \textit{feature space} as realized, for instance, by a neural network.
The transformations $\psi: \mathbb{R}^d \rightarrow \mathcal{P}$ and $\varphi: \mathcal{C} \rightarrow \mathcal{P}$ embed features and classes into a common \textit{prediction space} $\mathcal{P}$, respectively.
One of the simplest class embeddings, for example, consists in mapping each class to a one-hot vector:
\begin{equation}
\label{eq:onehot-embedding}
\setlength{\belowdisplayskip}{14pt}
\varphi_\mathrm{onehot}(y) =
\begin{bmatrix}
\smash[b]{\underbrace{0 \; \cdots \; 0}_{y-1\mathrm{\ times}}}
& 1 &
\smash[b]{\underbrace{0 \; \cdots \; 0}_{n-y\mathrm{\ times}}} \\
\end{bmatrix}^\top \;.
\end{equation}

We consider the class embeddings $\varphi$ as fixed and aim at learning the parameters $\theta$ of a neural network $f_\theta$ by maximizing the cosine similarity between the image features and the embeddings of their classes.
To this end, we define the \textit{cosine loss function} to be minimized by the neural network:
\begin{equation}
\label{eq:cosine-loss}
\cosloss(x, y) = 1 - \cossim\bigl(f_\theta(x), \varphi(y)\bigr) \;.
\end{equation}

In practice, this is implemented as a sequence of two operations.
First, the features learned by the network (with $d=n$) are $L^2$-normalized: $\psi(x) = \frac{x}{\|x\|_2}$.
This restricts the prediction space to the unit hypersphere, where the cosine similarity is equivalent to the dot product:
\begin{equation}
\label{eq:dot-prod-loss}
\cosloss(x,y) = 1 - \bigl\langle \varphi(y),\, \psi(f_\theta(x)) \bigr\rangle \;.
\end{equation}
The class embeddings $\varphi(y)$ need to lie on the unit hyper-sphere as well for this equation to hold. One-hot vectors, for example, have unit-norm by definition and hence do not need to be $L^2$-normalized explicitly.

When working with batches of multiple samples, we compute the average loss over all instances in the batch.

\subsection{Comparison with Categorical Cross-Entropy and Mean Squared Error}
\label{subsec:crossent-loss}

\noindent
In the following, we discuss the differences between the cosine loss and two other well-known loss functions: the categorical cross-entropy and the mean squared error (MSE).
The main difference between these loss functions lies in the type of the prediction space they assume, which determines how the dissimilarity between predictions and ground-truth labels is measured.

MSE is the simplest of these loss functions, since it does not apply any transformation to the feature space and hence uses an Euclidean prediction space.
Naturally, the dissimilarity between the samples and their classes in this space is measured by the squared Euclidean distance:
\begin{equation}
\label{eq:mse}
\mseloss(x, y) = \| f_\theta(x) - \varphi(y) \|_2^2 \;.
\end{equation}

The cosine loss introduced in the previous section restricts the prediction space to the unit hypersphere through $L^2$ normalization applied to the feature space.
In the resulting space, the squared Euclidean distance is equivalent to $\cosloss$ as defined in \eqref{eq:dot-prod-loss}, up to multiplication with a constant.

Categorical cross-entropy is the most commonly used loss function for learning a neural classifier.
As a proxy for the Kullback-Leibler divergence, it is a dissimilarity measure in the space of probability distributions, though it is not symmetric.
The \textit{softmax} activation is applied to transform the network output into this prediction space, interpreting it as the log-odds of the probability distribution over the classes.
The cross-entropy between the predicted and the true class distribution is then employed as dissimilarity measure:
\begin{equation}
\label{eq:crossent}
\xentloss(x, y) = -\left\langle \varphi(y),\, \log\mathopen{}\left( \frac{\exp(f_\theta(x))}{\|\exp(f_\theta(x))\|_1} \right)\mathclose{} \right\rangle \;,
\end{equation}
where $\exp$ and $\log$ are applied element-wise.

A comparison of these loss functions in a 2-dimensional feature space for a ground-truth class $y=1$ is shown in \cref{fig:loss-functions}.
Compared with cross-entropy and MSE, the cosine loss exhibits some distinctive properties:
First, it is bounded in $[0,2]$, while cross-entropy and MSE can take arbitrarily high values.
Secondly, it is invariant against scaling of the feature space, since it depends only on the direction of the feature vectors, not on their magnitude.

The cross-entropy loss function, in contrast, exhibits an area of steep descent and two widespread areas of comparatively small variations.
Note that the bright and dark regions in \cref{subfig:xent-loss} are not constant but the differences are just too small for being visible.
This makes the choice of a suitable initialization and learning rate schedule nontrivial.
In contrast, we expect the cosine loss to behave more robustly across different datasets with varying numbers of classes.

Furthermore, the optimum value of the cross-entropy loss function is obtained when the feature value for the dimension corresponding to the true class is much larger than that for any other class and approaches infinity \cite{szegedy2016rethinking,he2018bag}.
This is suspected to result in overfitting, which is a particularly important problem when learning from small datasets.
To mitigate this issue, \textit{label smoothing} \cite{szegedy2016rethinking} adds noise to the ground-truth distribution as regularization:
Instead of projecting the class labels onto one-hot vectors, the true class receives a probability of $1-\varepsilon$ and all remaining classes are assigned $\frac{\varepsilon}{n-1}$, where $\varepsilon$ is a small constant (e.g., $0.1$).
This makes the optimal network outputs finite and has been found to improve generalization slightly.

With respect to the cosine loss, on the other hand, the $L^2$ normalization serves as a regularizer, without the need for an additional hyper-parameter that would need to be tuned for each dataset.
Furthermore, there is not only one finite optimal network output, but an entire sub-space of optimal values.
This allows the training procedure to focus solely on the direction of feature vectors without being confused by Euclidean distance measures, which are problematic in high-dimensional spaces \cite{beyer1999nearest}.
Especially in face of small datasets, we assume that this invariance against scaling of the network output is a useful regularizer.

\subsection{Semantic Class Embeddings}
\label{subsec:semantic-embeddings}

\noindent
So far, we have only considered one-hot vectors as class embeddings, distributing the classes evenly across the feature space.
However, this ignores semantic relationships among classes, since some classes are more similar to each other than to other classes.
To overcome this, Barz and Denzler \cite{Barz19:Embeddings} proposed to derive class embeddings $\varphi_\mathrm{sem}$ on the unit hypersphere whose dot products equal the semantic similarity of the classes.
The measure for this similarity is derived from an ontology such as WordNet \cite{fellbaum1998wordnet}, encoding prior knowledge about the relationships among classes.
They then train a CNN to map images into this semantic feature space using the cosine loss.

They have shown that the integration of this semantic information improves the semantic consistency of content-based image retrieval results significantly.
With regard to classification accuracy, however, this method was only competitive with categorical cross-entropy when this was added as an additional loss function:
\begin{equation}
\begin{split}
    \label{eq:embed-cls-loss}
    \cosclsloss(x, y) = 1 - \bigl\langle \psi(f_\theta(x)),\, \varphi_\mathrm{sem}(y) \bigr\rangle \\
                          - \lambda \cdot \bigl\langle \varphi_\mathrm{onehot}(y),\, \log(g_\theta(\psi(f_\theta(x)))) \bigr\rangle \;,
\end{split}
\end{equation}
where $\lambda \in \mathbb{R}^+$ is a hyper-parameter and the transformation $g_\theta: \mathbb{R}^d \rightarrow \mathbb{R}^n$ is realized by an additional fully-connected layer with softmax activation.

Besides two larger ones, Barz and Denzler \cite{Barz19:Embeddings} also analyzed one small dataset.
This was the only case where their method also provided superior classification accuracy than categorical cross-entropy.
In the following, we apply the cosine loss with and without semantic embeddings to several small datasets and show that the largest part of this effect is actually not due to the prior knowledge but the cosine loss.

\section{Datasets}
\label{sec:datasets}

\begin{table}[t]
    \small
    \setlength{\tabcolsep}{4pt}\centering
    \begin{tabularx}{\linewidth}{Xrrrr}
        \toprule
        Dataset & \#Classes & \#Training & \#Test & Samples/Class \\
        \midrule
        CUB & 200 & 5,994 & 5,794 & 29 -- 30 (30) \\
        NAB & 555 & 23,929 & 24,633 & 4 -- 60 (44) \\
        Cars & 196 & 8,144 & 8,041 & 24 -- 68 (42) \\
        Flowers-102 & 102 & 2,040 & 6,149 & 20 \\
        MIT Indoor & 67 & 5,360 & 1,340 & 77 -- 83 (80) \\
        CIFAR-100 & 100 & 50,000 & 10,000 & 500 \\
        \bottomrule
    \end{tabularx}
    \caption{Image dataset statistics. The number of samples per class refers to training samples and numbers in parentheses specify the median.}
    \label{tbl:dataset-statistics}
\end{table}

\noindent
We conduct experiments on five small image datasets as well as a larger one.
Statistics for all datasets can be found in \cref{tbl:dataset-statistics}.
Moreover, we show results on a text classification task to demonstrate that the benefit of the cosine loss is not exclusive to the image domain.

\subsection{CUB and NAB}
\label{subsec:cub-nab}

\noindent
The Caltech-UCSD Birds-200-2011 (CUB) \cite{WahCUB_200_2011} and the North American Birds (NAB) \cite{vanhorn2015nab} datasets are quite similar.
Both are fine-grained datasets of bird species, but NAB comprises four times more images than CUB and almost three times more classes.
It is also even more fine-grained than CUB and distinguishes between male/female and adult/juvenile birds of the same species.

In contrast to CUB, the NAB dataset already provides a hierarchy of classes.
To enable experiments with semantic class embeddings on CUB as well, we created a hierarchy for this dataset manually.
We used information about the scientific taxonomy of bird species that is available in the Wikispecies project \cite{WikiSpecies}.
This resulted in a hierarchy where the 200 bird species of CUB are identified by their scientific names and organized by order, sub-order, super-family, family, sub-family, and genus.

While order, family, and genus exist in all branches of the hierarchy, sub-order, super-family, and sub-family are only available for some of them.
This leads to an unbalanced taxonomy tree where not all species are at the same depth.
To overcome this issue and analyze the effect of the depth of the hierarchy on classification accuracy, we derived two balanced variants:
a flat one consisting only of the order, family, genus, and species level, and a deeper one comprising 7 levels.
For the latter one, we manually searched for additional information about missing super-orders, sub-orders, super-families, sub-families, and tribes in the English Wikipedia \cite{Wikipedia} and The Open Tree of Life \cite{OpenTreeOfLife}.

Since CUB is a very popular dataset in the fine-grained community and other projects could benefit from this class hierarchy as well, we make it available at \url{https://github.com/cvjena/semantic-embeddings/tree/v1.2.0/CUB-Hierarchy}.

\begin{table*}[t]
    \small
    \begin{tabularx}{\linewidth}{Xcccccc}
        \toprule
        & CUB & NAB & Cars & Flowers-102 & MIT Indoor & CIFAR-100 \\
        \midrule
        MSE & 42.0 & 27.7 & 41.8 & 63.0 & 38.2 & 75.1 \\
        softmax + cross-entropy & 51.9 & 59.4 & 78.2 & 67.3 & 44.3 & 77.0 \\
        softmax + cross-entropy + label smoothing & 55.5 & 68.3 & 78.1 & 66.8 & 38.7 & \textbf{77.5} \\
        cosine loss (one-hot embeddings) & 67.6 & 71.7 & 84.3 & \textbf{71.1} & 51.5 & 75.3 \\
        cosine loss + cross-entropy (one-hot embeddings) & \textbf{68.0} & \textbf{71.9} & \textbf{85.0} & 70.6 & \textbf{52.7} & 76.4 \\
        \midrule
        cosine loss (semantic embeddings) & 59.6 & 72.1 & --- & --- & --- & 74.6 \\
        cosine loss + cross-entropy (semantic embeddings) & 70.4 & 73.8 & --- & --- & --- & 76.7 \\
        \midrule
        fine-tuned softmax + cross-entropy & 82.5 & 80.1 & 91.2 & 97.2 & 79.9 & --- \\
        fine-tuned cosine loss (one-hot embeddings) & 82.7 & 78.6 & 89.6 & 96.2 & 74.3 & --- \\
        fine-tuned cosine loss + cross-entropy (one-hot embeddings) & 82.7 & 81.2 & 90.9 & 96.2 & 73.3 & --- \\
        \bottomrule
    \end{tabularx}
    \caption{Test-set classification accuracy in percent (\%) achieved with different loss functions on various datasets. The best value per column not using external data or information is set in bold font.}
    \label{tbl:performance}
\end{table*}

\subsection{Cars, Flowers-102, and MIT Indoor Scenes}
\label{subsec:cars-flowers}

\noindent
The Stanford Cars \cite{krause2013cars} and Oxford Flowers-102 \cite{nilsback2008flowers} datasets are two well-known fine-grained datasets of car models and flowers, respectively.
They are not particularly challenging anymore nowadays, but we include them in our experiments to avoid a bias towards bird recognition.
To furthermore also include a dataset from the pre-deep-learning era that is not from the fine-grained recognition domain, we also conduct experiments on the MIT 67 Indoor Scenes dataset \cite{quattoni2009mitscenes}, which contains images of 67 different indoor scenes.
All three datasets do not provide a class hierarchy and we will hence only conduct experiments on them in combination with one-hot class embeddings.

\subsection{CIFAR-100}
\label{subsec:cifar-100}

\noindent
With 500 training images per class, the CIFAR-100 \cite{krizhevsky2009cifar} dataset does not fit into our definition of a \textit{small dataset} from \cref{sec:introduction}.
However, we can sub-sample it to interpolate between small and large datasets for quantifying the effect of the number of samples per class on the performance of categorical cross-entropy and the cosine loss.

A hierarchy for the classes of the CIFAR-100 dataset derived from WordNet \cite{fellbaum1998wordnet} has recently been provided in \cite{Barz19:Embeddings}.
We use this taxonomy in our experiments with semantic class embeddings.

\subsection{AG News}
\label{subsec:ag-news}

\noindent
To demonstrate that the benefits of the cosine loss are not exclusive to image datasets, we also conduct experiments on the widely used variant of the AG News text classification dataset introduced in \cite{zhang2015character}.
This is a large-scale dataset comprising the titles and descriptions of 120,000 training and 7,600 validation news articles from 4 categories.
For our experiments, we will sub-sample the dataset to assess the difference between the cosine loss and cross-entropy for text datasets of different size.

\section{Experiments}
\label{sec:experiments}

\noindent
To demonstrate the performance of the cosine loss on the aforementioned small datasets, we compare it with the categorical cross-entropy loss and MSE.
Moreover, we analyze the effect of the dataset size and prior semantic knowledge.

\subsection{Setup}
\label{subsec:setup}

\noindent
For CIFAR-100, we train a ResNet-110 \cite{he2016resnet} with an input image size of $32 \times 32$ and twice the number of channels per layer, as suggested by \cite{Barz19:Embeddings}.
For all other image datasets, we use a standard ResNet-50 architecture \cite{he2016resnet}.
Images from the four fine-grained datasets are resized so that their smaller side is $512$ pixels wide and randomly cropped to $448 \times 448$ pixels.
For MIT Indoor Scenes, we resize images to 256 pixels and use $224 \times 224$ crops.
We furthermore apply random horizontal flipping and random erasing \cite{zhong2017random}.

For training the network, we follow the learning rate schedule of \cite{Barz18:GoodTraining}:
5 cycles of Stochastic Gradient Descent with Warm Restarts (SGDR) \cite{loshchilov2016sgdr} with a base cycle length of 12 epochs.
The number of epochs is doubled at the end of each cycle, amounting to a total number of 372 epochs.
During each epoch, the learning rate is smoothly reduced from a pre-defined maximum learning rate $\mathrm{lr}_\mathrm{max}$ down to $10^{-6}$ using cosine annealing.
To prevent divergence caused by initially high learning rates, we employ gradient clipping \cite{pascanu2013gradientclipping} to a maximum norm of $10$.
For CIFAR-100, we perform training on a single GPU with 100 images per batch.
For all other datasets, we distribute a batch of 96 samples (256 samples for MIT Indoor Scenes) across 4 GPUs.

For text classification on AG News, we use GloVe word embeddings \cite{pennington2014glove} pre-trained on Wikipedia to represent each word as 300-dimensional vector and train a GRU layer \cite{cho2014properties} with 300 units and a dropout ratio of 0.5 for both input and output, followed by batch-normalization and a fully-connected layer for classification.
The same learning rate schedule as for the vision experiments is applied, but limited to 180 epochs using a batch size of 128 samples.

\subsection{Performance Comparison}
\label{subsec:performance}

\noindent
First, we examine the performance obtained by training with the cosine loss and investigate the additional use of prior semantic knowledge.
Therefore, we report the classification accuracy of the cosine loss with semantic embeddings and with one-hot embeddings in \cref{tbl:performance} and compare it with the performance of MSE and standard softmax with cross-entropy.
For the CUB dataset, we have used the deep variant of the class hierarchy for this experiment.
Other variants will be analyzed in \cref{subsec:semantic-information}.

With regard to cross-entropy, we also examine the use of label smoothing (cf.\ \cref{subsec:crossent-loss}).
We set its hyper-parameter to $\varepsilon=0.1$, as suggested by Szegedy et al.~\cite{szegedy2016rethinking}.

As an upper bound, we also report the classification accuracy achieved by fine-tuning a network pre-trained on ILSVRC'12 \cite{russakovsky2015ilsvrc}, using the weights from He et al.\ \cite{he2016resnet}.

Regarding the cosine loss, we report the performance of two variants: the cosine loss alone as in \eqref{eq:dot-prod-loss} and combined with cross-entropy after an additional fully-connected layer as in \eqref{eq:embed-cls-loss}.
In the latter case, we fixed the combination hyper-parameter $\lambda = 0.1$, following Barz and Denzler \cite{Barz19:Embeddings}.

To avoid any bias in favor of a certain method due to the maximum learning rate $\mathrm{lr}_\mathrm{max}$, we fine-tuned it for each method individually by reporting the best results from the set $\mathrm{lr}_\mathrm{max} \in \{2.5, 1.0, 0.5, 0.1, 0.05, 0.01, 0.005, 0.001\}$.
Since overfitting also began at different epochs for different loss functions, we do not report the final performance after all 372 epochs in \cref{tbl:performance}, but the best performance achieved after any epoch.

\paragraph{Results}

It can be seen that the classification accuracy obtained with the cosine loss outperforms cross-entropy after softmax considerably on all small datasets, with the largest relative improvements being 30\% and 21\% on the CUB and NAB dataset.
On Cars and Flowers-102, the relative improvements are 8\% and 6\%, but these datasets are easier in general.
The label smoothing technique \cite{szegedy2016rethinking}, on the other hand, leads to an improvement on CUB and NAB only and still falls behind the cosine loss by a large margin.
When a sufficiently large dataset such as CIFAR-100 is used, however, cross-entropy and the cosine loss perform similarly well. MSE performs very poorly on most datasets and never better than any of the other loss functions.

Before the rise of deep learning and pre-training, leading methods for fine-grained recognition achieved an accuracy of 57.8\% on CUB by making use of the object part annotations provided for the training set \cite{Goering14:NPT}.
While this approach performs better than training a CNN with cross-entropy on this small dataset, it is outperformed by the cosine loss, even without the use of additional information.

Still, there is a large gap between training from scratch and fine-tuning a network pre-trained on a million of images from ImageNet.
Our results show, however, that this gap can in fact be reduced.

\subsection{Effect of Semantic Information}
\label{subsec:semantic-information}

\noindent
Besides one-hot vectors as class embeddings, we also experimented with hierarchy-based semantic embeddings \cite{Barz19:Embeddings}.
The results in \cref{tbl:performance} show a slight increase in performance by one percent point on NAB and 3 percent points on CUB, but this difference is rather small compared to the 17 percent points improvement over cross-entropy on CUB achieved by the cosine loss alone.

To analyze the influence of semantic information derived from class taxonomies further, we experimented with three hierarchy variants of different depth on the CUB dataset (cf.\ \cref{subsec:cub-nab}).
\Cref{tbl:cub-hierarchy-comparison} shows the classification performance obtained with the cosine loss for each of the hierarchies.

When using one-hot embeddings, the difference between the cosine loss alone ($\cosloss$) and the cosine loss combined with cross-entropy ($\cosclsloss$) is smallest.
When the class hierarchy grows deeper, however, the performance of $\cosloss$ decreases, while the classification accuracy achieved by $\cosclsloss$ improves.

We attribute this to the fact that semantic embeddings do not enforce the separability of all classes to the same extent.
With one-hot embeddings, all classes are equally far apart.
The aim of hierarchy-based semantic embeddings, however, is to place semantically similar classes close together in the feature space and dissimilar classes far apart.
Since the cosine loss corresponds to the distance of samples from their class center in that semantic space, confusing similar classes is not penalized as much as confusing dissimilar classes.
This is why the additional integration of cross-entropy improves accuracy in such a scenario by enforcing the separation of all classes homogeneously.

\subsection{Effect of Dataset Size}
\label{subsec:dataset-size}

\begin{figure*}[t]
    \includegraphics[width=\linewidth]{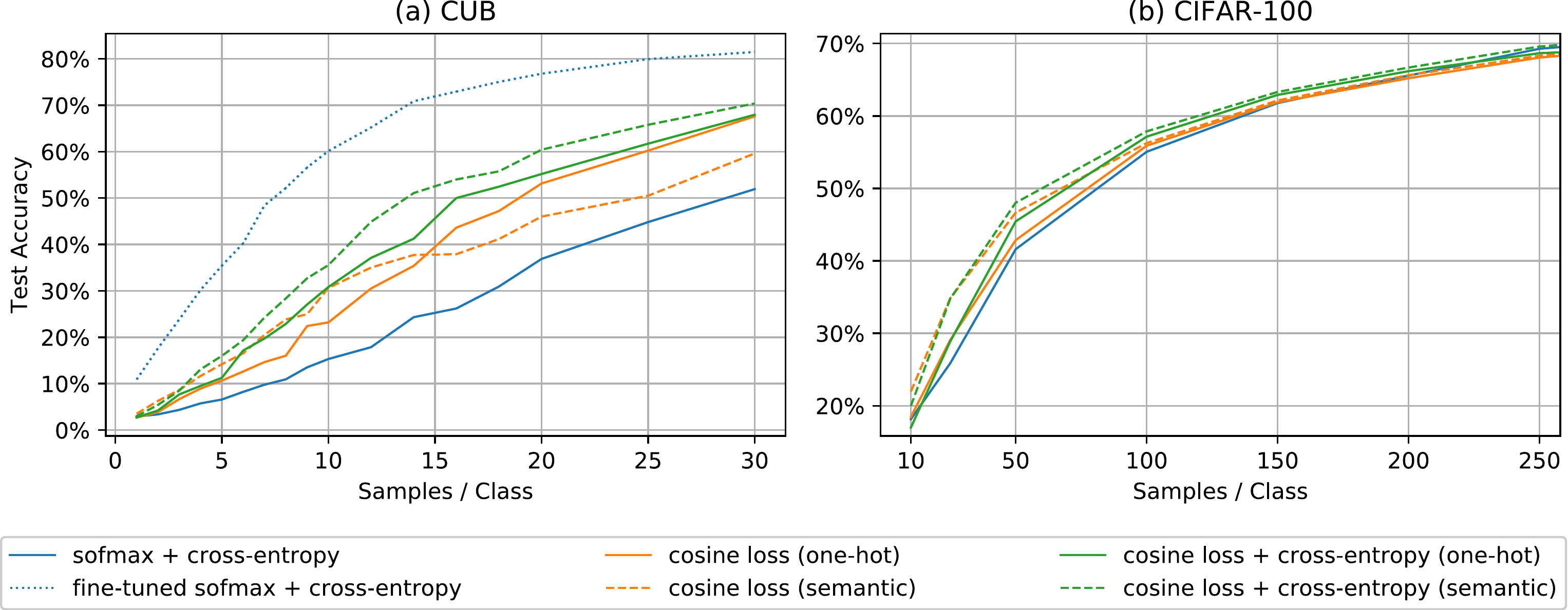}%
    {\phantomsubcaption\label{subfig:cub-size}}%
    {\phantomsubcaption\label{subfig:cifar-size}}%
    \caption{Classification performance depending on the dataset size.}
    \label{fig:dataset-size}
\end{figure*}

\noindent
To examine the behavior of the cosine loss and the cross-entropy loss depending on the size of the training dataset, we conduct experiments on several sub-sampled versions of CUB and CIFAR-100.
We specify the size of a dataset in the number of samples per class and vary this number from 1 to 30 for CUB, while we use 10, 25, 50, 100, 150, 200, and 250 samples per class for CIFAR-100.
For each experiment, we choose the respective number of samples from each class at random and increase the number of iterations per epoch, so that the total number of iterations is approximately constant.
The performance is always evaluated on the full test set.
For CIFAR-100, we report the mean over 3 runs.
To facilitate comparability between experiments with different dataset sizes, we fixed $\mathrm{lr}_\mathrm{max}$ to the best value identified for each method individually in \cref{subsec:performance}.

The results depicted in \cref{fig:dataset-size} emphasize the benefits of the cosine loss for learning from small datasets.
On CUB, the cosine loss results in consistently better classification accuracy than the cross-entropy loss and also improves faster when more samples are added.
Including semantic information about the relationships among classes seems to be most helpful in scenarios with very few samples.
The same holds true for the combination of the cosine loss and the cross-entropy loss, but since this performed slightly better than the cosine loss alone in all cases, we would recommend this variant for practical use in general.

Nevertheless, all methods are still largely outperformed on CUB by pre-training on ILSVRC'12.
This is barely a surprise, since the network has seen 200 times more images in this case.
We have argued in \cref{sec:introduction} why this kind of transfer learning can sometimes be problematic (e.g., domain shift, legal restrictions).
In such a scenario, better methods for learning from scratch on small datasets, such as the cosine loss proposed here, are crucial.

The experiments on CIFAR-100 allow us to smoothly transition from small to larger datasets.
The gap between the cosine loss and cross-entropy is smaller here, but still noticeable and consistent.
It can also be seen that cross-entropy starts to take over from 150--200 samples per class.

\begin{table}[t]
    \begin{tabularx}{\linewidth}{Xccc}
        \toprule
        Embedding & Levels & $\cosloss$ & $\cosclsloss$ \\
        \midrule
        one-hot & 1 & \textbf{67.6} & 68.0 \\
        flat & 4 & 66.6 & 68.8 \\
        Wikispecies & 4-6 & 61.6 & 69.9 \\
        deep & 7 & 59.9 & \textbf{70.4} \\
        \bottomrule
    \end{tabularx}
    \caption{Accuracy in \% on the CUB test set obtained by cosine loss with class embeddings derived from taxonomies of varying depth. The best value per column is set in bold.}
    \label{tbl:cub-hierarchy-comparison}
\end{table}

\subsection{Results for text classification}
\label{subsec:text-classification}

\begin{figure}[t]
    \includegraphics[width=\linewidth]{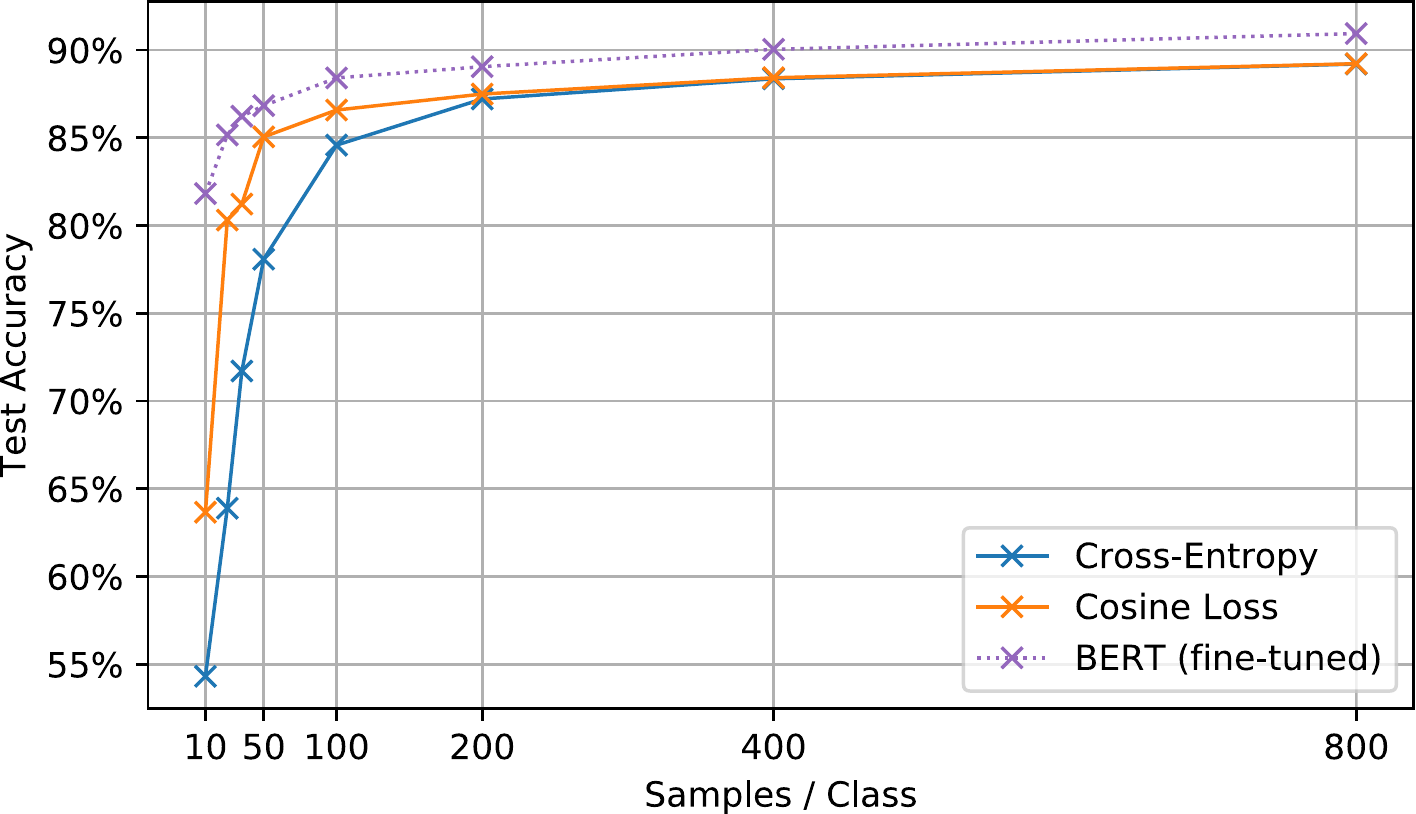}
    \caption{Validation accuracy achieved using the cross-entropy and the cosine loss on sub-sampled versions of the AG News dataset, averaged over 10 runs.}
    \label{fig:agnews-performance}
\end{figure}

We conduct a similar analysis on the AG News text classification dataset by sub-sampling it and training on sub-sets of 10, 25, 35, 50, 100, 200, 400, and 800 samples per class.
Each experiment is repeated ten times on different random subsets of the data and we report the average of the maximum validation accuracy achieved during training in \cref{fig:agnews-performance}.
As an upper bound, we also show the accuracy that can be obtained by fine-tuning a BERT model \cite{devlin2019bert} pre-trained on huge text corpora.
Note that this is not a fair comparison, since BERT uses a complex transformer architecture with 10 times more parameters than our simple GRU model.

It can be seen that the cosine loss achieves substantially better performance than cross-entropy for datasets with up to 100 documents per class.
This is in accordance with our findings on the CUB dataset.
For the smallest test cases with only 10 and 25 samples per class, the relative improvement of cosine loss over cross-entropy is 17\% and 26\%, respectively.
To assess whether the difference of the means over the ten runs are statistically significant, we have conducted Welch's t-test. For up to 100 samples per class, the differences are highly significant with p-Values less than 1\%, while there is no significant difference between categorical cross-entropy and cosine loss for larger datasets.

\section{Conclusions}
\label{sec:conclusions}

\noindent
We have found the cosine loss to be useful for training deep neural classifiers from scratch on limited data.
Experiments on five well-known small image datasets and one text classification task have shown that this loss function outperforms the traditionally used cross-entropy loss after softmax activation by a large margin.
On the other hand, both loss functions perform similarly if a sufficient amount of training data is available or the network is initialized with weights pre-trained on a large dataset.

This leads to the hypothesis, that the $L^2$ normalization involved in the cosine loss is a strong regularizer.
Evidence for this hypothesis is provided by the poor performance of the MSE loss, which mainly differs from the cosine loss by not applying $L^2$ normalization.
Previous works have found that direction bears substantially more information in high-dimensional feature spaces than magnitude \cite{husain2017rvd,zhe2018directional}.
The magnitude of feature vectors can hence mainly be considered as noise, which is eliminated by $L^2$ normalization.
Moreover, the cosine loss is bounded between 0 and 2, which facilitates a dataset-independent choice of a learning rate schedule and limits the impact of misclassified samples, e.g., difficult examples or label noise.

We have furthermore analyzed the effect of the dataset size by performing experiments on sub-sampled variants of two image and one text classification dataset and found the cosine loss to perform better than cross entropy for datasets with less than 200 samples per class.

Moreover, we investigated the benefit of using semantic class embeddings instead of one-hot vectors as target values.
While doing so did result in a higher classification accuracy, the improvement was rather small compared to the large gain caused by the cosine loss itself.

While some problems can in fact be solved satisfactorily by simply collecting more and more data, we hope that applications that have to deal with limited amounts of data and cannot apply pre-training can benefit from using the cosine loss.
Moreover, we hope to motivate future research on different loss functions for classification, since there obviously are viable alternatives to categorical cross-entropy.

\subsubsection*{Acknowledgements}

\begin{footnotesize}
This work was supported by the German Research Foundation as part of the
priority programme ``Volunteered Geographic Information: Interpretation,
Visualisation and Social Computing'' (SPP 1894, contract number DE 735/11-1).
We also gratefully acknowledge the support of NVIDIA Corporation with the
donation of Titan Xp GPUs used for this research.
\par
\end{footnotesize}

{\small
    \bibliographystyle{ieee}
    \bibliography{references}
}

\end{document}